\documentclass[lettersize,journal]{IEEEtran}
\usepackage{amsmath,amsfonts}
\usepackage{algorithm}
\usepackage{array}
\usepackage[caption=false,font=normalsize,labelfont=sf,textfont=sf]{subfig}
\usepackage{textcomp}
\usepackage{stfloats}
\usepackage{url}
\usepackage{verbatim}
\usepackage{graphicx}
\usepackage{cite}
\usepackage{hyperref}
\hypersetup{
colorlinks = true, 
urlcolor = magenta, 
linkcolor = red, 
citecolor = blue}
\usepackage{enumitem}
\usepackage{romannum}
\usepackage{booktabs}
\usepackage{amsmath}
\usepackage{cuted}
\usepackage{capt-of}

\usepackage{array}
\newcolumntype{P}[1]{>{\centering\arraybackslash}p{#1}}
\usepackage{algorithm}
\usepackage{algpseudocode}
\usepackage{multirow}

\usepackage{pythonhighlight}

\begin{document}

\title{Treating Motion as Option with Output Selection\\for Unsupervised Video Object Segmentation}

\author{
\vspace{2mm}
Suhwan Cho\quad Minhyeok Lee\quad Jungho Lee\quad MyeongAh Cho\\
Seungwook Park\quad Jaeyeob Kim\quad Hyunsung Jang\quad Sangyoun Lee,\IEEEmembership{ Member, IEEE}

\thanks{
This work is extended from the preliminary exploration, ``Treating Motion as Option to Reduce Motion Dependency in Unsupervised Video Object Segmentation", presented in \textit{Proceedings of the IEEE/CVF Winter Conference on Applications of Computer Vision}, pp. 5140-5149, Jan. 2023.
\vspace{1mm}

Suhwan Cho, Minhyeok Lee, Jungho Lee, and Sangyoun Lee are with the School of Electrical and Electronic Engineering, Yonsei University, Seoul, Korea (e-mail: \{chosuhwan, hydragon516, 2015142131, syleee\}@yonsei.ac.kr). MyeongAh Cho is with the Department of Software Convergence, Kyung Hee University, Yongin, Korea (e-mail: maycho@khu.ac.kr). Seungwook Park, Jaeyeob Kim, and Hyunsung Jang are with the EO/IR Systems R\&D Lab, LIG Nex1, Yongin, Korea (e-mail: \{seungwook.park, jaeyeob.kim, hyunsung.jang\}@lignex1.com). This work was supported by KRIT-Grant funded by Defense Acquisition Program Administration (DAPA) (20-102-EOO-013). 

\vspace{1mm}
Copyright © 2025 IEEE. Personal use of this material is permitted. However, permission to use this material for any other purposes must be obtained from the IEEE by sending an email to pubs-permissions@ieee.org.
}}

\maketitle

\begin{strip}
\vspace{-3.9cm}
\centering
\includegraphics[width=1.0\textwidth]{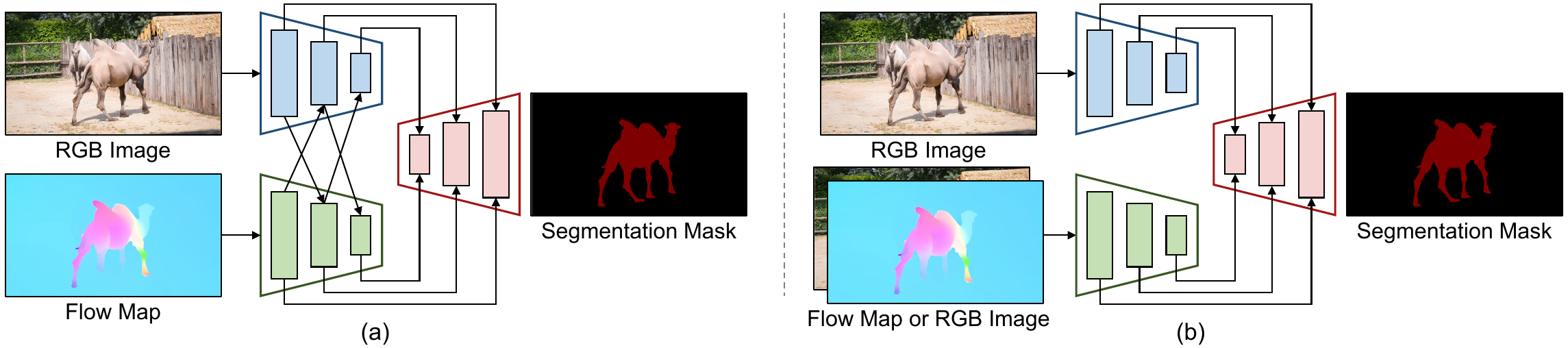}
\vspace{-6mm}
\captionof{figure}{Visualized comparison of (a) a conventional two-stream VOS network and (b) our proposed motion-as-option network. Unlike existing methods, our motion-as-option network is designed to handle both RGB images and optical flow maps as motion inputs.\label{figure1}}
\end{strip}

\begin{abstract}
Unsupervised video object segmentation aims to detect the most salient object in a video without any external guidance regarding the object. Salient objects often exhibit distinctive movements compared to the background, and recent methods leverage this by combining motion cues from optical flow maps with appearance cues from RGB images. However, because optical flow maps are often closely correlated with segmentation masks, networks can become overly dependent on motion cues during training, leading to vulnerability when faced with confusing motion cues and resulting in unstable predictions. To address this challenge, we propose a novel motion-as-option network that treats motion cues as an optional component rather than a necessity. During training, we randomly input RGB images into the motion encoder instead of optical flow maps, which implicitly reduces the network's reliance on motion cues. This design ensures that the motion encoder is capable of processing both RGB images and optical flow maps, leading to two distinct predictions depending on the type of input provided. To make the most of this flexibility, we introduce an adaptive output selection algorithm that determines the optimal prediction during testing. Code and models are available at \url{https://github.com/suhwan-cho/TMO}.
\end{abstract}


\section{Introduction}
\IEEEPARstart{W}{ith} the rapid advancements in deep learning technologies, video processing techniques have made remarkable progress. One of the most notable advancements is in video segmentation. Recent tasks in video segmentation include semantic separation of videos~\cite{vss1, vss2, uss}, object segmentation based on mask guidance~\cite{vos1, vos2, vos3, prm} or language guidance~\cite{vos4}, and the automatic detection of primary objects~\cite{vos5, vos6, vos7, vos8, vos9} within video sequences. In this context, we focus on unsupervised video object segmentation (VOS), also known as video salient object detection (SOD). This task aims to detect and segment the most salient objects at the pixel level without any external guidance. Without explicit object guidance, the challenge lies in automatically identifying and consistently segmenting these primary objects throughout the video sequence. While termed ``unsupervised", this task does not relate to unsupervised learning protocols; rather, it refers to the fact that the target object is not explicitly guided but instead must be identified autonomously by the model based on inherent saliency present within the video data.

Salient objects often exhibit distinct movements compared to the background. Recent approaches to unsupervised VOS utilize motion cues extracted from optical flow maps in conjunction with appearance cues from RGB images. These features are fused during the embedding process with mutual guidance, as illustrated in Figure~\ref{figure1} (a). For example, MATNet~\cite{MATNet} employs a two-stream interleaved encoder to enhance appearance features with attentive motion features. FSNet~\cite{FSNet} introduces a relational cross-attention module to extract discriminative features from both appearance and motion cues, ensuring mutual restraint between them. AMC-Net~\cite{AMC-Net} uses a co-attention gate module in the bilateral encoder branches to balance the contributions of multi-modal features. HFAN~\cite{HFAN} proposes a hierarchical mechanism to align and fuse appearance and motion features. However, these methods are susceptible to the quality of optical flow maps, as their network structures heavily depend on motion cues.

To address this limitation, we propose a novel motion-as-option network with a simple encoder-decoder architecture that operates regardless of the availability of motion cues, as shown in Figure~\ref{figure1}~(b). Our network includes two encoders that extract semantic features from appearance and motion inputs, respectively, while the decoder integrates these embedded features to produce an object segmentation mask. Unlike existing two-stream VOS approaches, our method randomly activates or deactivates the motion stream during training. When activated, optical flow maps are used as motion input, incorporating both appearance and motion cues for mask prediction. When deactivated, RGB images are used as motion input, relying solely on appearance cues for prediction. By not always providing motion cues (which can be overly dominant) during training, the network becomes less dependent on the motion encoder, significantly enhancing robustness against confusing optical flow maps during inference.

To train the motion-as-option network effectively, we prepare training samples with and without optical flow maps. While a straightforward approach is to discard optical flow maps in some VOS training samples, we further enhance our training by incorporating SOD samples. When using VOS samples, optical flow maps are fed into the motion encoder, while SOD samples use RGB images. This collaborative learning strategy leverages a larger dataset to maximize the network's flexibility. After training, the motion-as-option network can handle both RGB images and optical flow maps as motion inputs. During inference, it can generate two predictions based on the input type. To select the optimal prediction, we propose an output selection algorithm that calculates and compares overall confidence scores for the two outputs, choosing the one with the higher score. This method proves more effective than conventional soft ensemble approaches for the motion-as-option network.

On public benchmark datasets for unsupervised VOS, including the DAVIS 2016~\cite{DAVIS} validation set, the FBMS~\cite{FBMS} test set, the YouTube-Objects~\cite{YTOBJ} dataset, and the Long-Videos~\cite{LVID} dataset, our approach achieves state-of-the-art performance while maintaining real-time inference speed. Extensive analysis confirms the effectiveness of each proposed component. Our solution is both simple and fast, offering robust performance that represents a significant advancement toward practical VOS systems. It also establishes a solid baseline for future research in the field.

This paper builds upon our previous conference work~\cite{TMO}, advancing the initial version in several notable aspects. First, we introduce the adaptive output selection algorithm to fully harness the unique capabilities of the motion-as-option network. Second, we significantly expand the experimental evaluation, offering more comprehensive validation through assessments on additional testbeds and backbone settings, detailed discussions of failure cases, and enhanced visualization results. Third, we provide an in-depth analysis of the motion-as-option network and the output selection protocol, thoroughly showcasing their strengths.

This work compares with several closely related methods published in the IEEE Transactions on Circuits and Systems for Video Technology, all of which aim to segment objects in videos. SSM-VOS~\cite{vos2} segments target objects designated in the first frame with binary masks by leveraging a separable structure modeling approach that captures structural information of the target during feature matching between the first and query frames. DMFormer~\cite{vos4} focuses on segmenting target objects designated by language guidance, employing a strategy to decouple input text into subject and context, facilitating explicit feature embedding for the target object. While these methods effectively segment objects within videos, they rely on explicit external guidance to designate target objects. In contrast, our method focuses on the automatic detection of primary objects within videos without external guidance.

FEM-Net~\cite{vos5} and IMCNet~\cite{vos6} focus on the automatic detection of primary objects, aligning more closely with our method. FEM-Net employs a flow edge connection module to estimate object boundaries from optical flow maps, using these boundaries as primary cues during feature embedding. IMCNet avoids explicit flow estimation by implicitly modeling motion cues through affinity computation between adjacent frames, reducing the reliance on potentially erroneous flow maps. Similar to our approach, these methods leverage motion cues to complement appearance cues, utilizing the distinctive movements of primary objects. However, unlike these methods, our approach explicitly addresses the issue of over-dependence on motion cues. Instead of introducing additional learnable modules, we propose a novel network training scheme to mitigate this dependency at a fundamental level. In addition, we introduce a selective inference strategy to enhance system stability, an aspect that has not been explored in previous work.

\begin{table*}[t] 
\centering 
\small 
\caption{Summary of the core concepts of related methods.} 
\vspace{-2mm}
\begin{tabular}{lP{1.7cm}l} 
\toprule
Method &Publication &Core Concept\\
\midrule 
MATNet~\cite{MATNet} &AAAI'20 &Uses asymmetric attention to modulate appearance features with motion features during encoding.\\
RTNet~\cite{RTNet} &CVPR'21 &Transforms appearance features to reduce ambiguity in motion features.\\
FSNet~\cite{FSNet} &ICCV'21 &Employs bi-directional message propagation between appearance and motion features.\\
TransportNet~\cite{TransportNet} &ICCV'21 &Applies the Sinkhorn algorithm to propagate features between modalities.\\
AMC-Net~\cite{AMC-Net} &ICCV'21 &Learns a gating mechanism to weigh the importance of appearance and motion features.\\
HFAN~\cite{HFAN} & ECCV'22 &Aligns appearance and motion features with primary object representations.\\
\midrule 
\textbf{TMO} & &Adjusts the training scheme to reduce reliance on the motion encoder.\\
\textbf{TMO++} & &Dynamically utilizes motion cues by running two independent scenarios.\\
\bottomrule
\end{tabular}
\label{table1} 
\end{table*}

\section{Related Work}
\noindent\textbf{Temporal coherence.} Video sequences typically exhibit similar visual content across successive frames, with salient objects remaining consistent. Early unsupervised VOS methods leveraged this temporal locality. COSNet~\cite{COSNet} introduces a global co-attention mechanism to emphasize regions with frequent coherence across frames. AGNN~\cite{AGNN} uses graph neural networks to represent frames as nodes in a fully connected graph, iteratively updating relationships through message passing. AD-Net~\cite{AD-Net} and F2Net~\cite{F2Net} address long-term dependencies by computing dense correspondences between pixels in reference and query frames. IMP~\cite{IMP} integrates an image-level SOD model with a semi-supervised VOS model for mask propagation. DPA~\cite{DPA} and GSA-Net~\cite{GSA-Net} use uniformly sampled reference frames to capture global video tendencies. However, these methods rely on multiple frames, complicating independent frame inference.

\vspace{1mm} 
\noindent\textbf{Motion information.} Optical flow maps are often used alongside RGB images in unsupervised VOS, as salient objects typically exhibit motion that contrasts with the background. MATNet~\cite{MATNet} introduces a motion-attentive transition block to asymmetrically transform appearance features into motion-attentive representations, enabling the appearance encoder to adaptively modulate feature embedding, unlike independent encoders. FSNet~\cite{FSNet} employs a bi-directional interaction module to fuse appearance and motion features using a full-duplex strategy, where features from each modality are modulated by signals generated from the other to ensure mutual constraint. AMC-Net~\cite{AMC-Net} utilizes a multi-modality co-attention module to assess the quality of features from each modality, using a gating function to balance contributions and suppress redundant or misleading information. Similarly, RTNet~\cite{RTNet} introduces a reciprocal transformation network to handle noisy flow maps by correlating intra-frame contrast, motion cues, and temporal coherence, while TransportNet~\cite{TransportNet} reduces noise through optimal structural matching with the Sinkhorn algorithm.

Existing two-stream methods primarily focus on integrating motion cues from optical flow maps with appearance cues. However, because optical flow maps are closely tied to segmentation masks, networks often become overly reliant on these cues, which can be unreliable during inference. While some methods attempt to address this issue by introducing modules to reduce the influence of noisy flows, these structural modifications applied after motion encoding fail to tackle the problem fundamentally. In contrast, our approach enhances the robustness of the motion encoder by directly refining the motion encoding process. The effectiveness of our method in handling noisy flows is further validated in Section~\ref{analysis}. A summarized comparison of closely related methods is presented in Table~\ref{table1}, highlighting our focus on external pipelines, in contrast to others' emphasis on internal modules.

\begin{figure*}[t]
\centering
\includegraphics[width=1\linewidth]{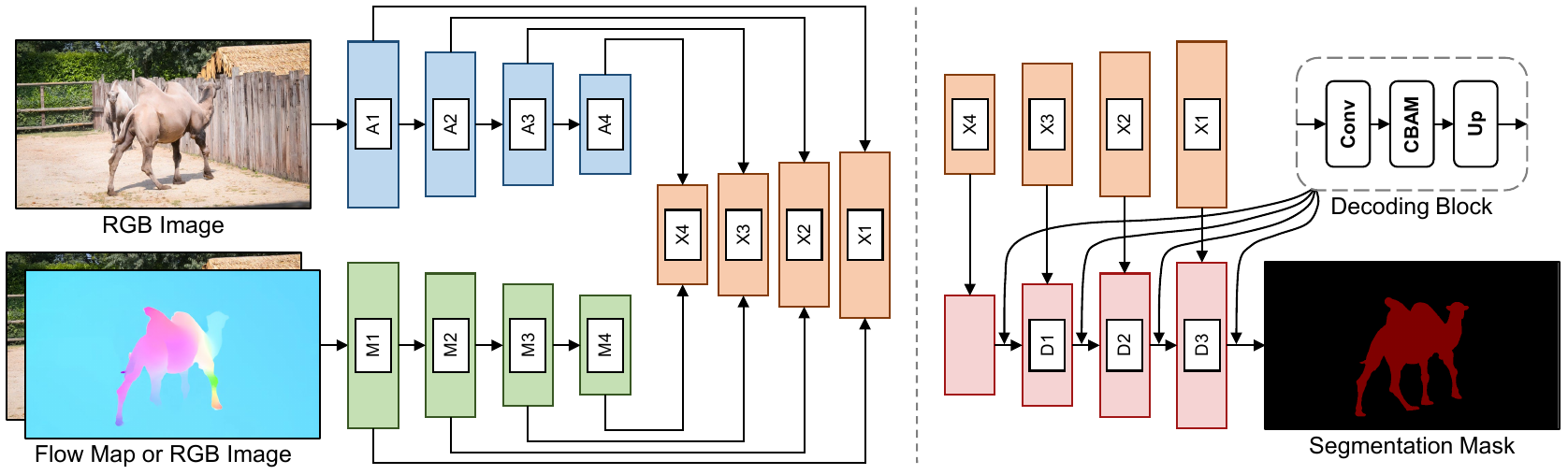}
\vspace{-6mm}
\caption{Architecture of our proposed network. When motion cues are leveraged, an optical flow map serves as the motion input. Alternatively, when motion cues are not utilized, an RGB image is used as the motion input. The network extracts appearance and motion features through separate encoders, which are then fused together and gradually decoded to produce the final segmentation mask.}
\label{figure2}
\end{figure*}

\section{Approach}
\subsection{Problem Formulation}
Given an input video sequence, an unsupervised VOS model aims to predict segmentation masks for the salient object. Following recent methods, we generate optical flow maps using a pre-trained optical flow estimation model and utilize these motion cues alongside appearance cues. Note that the 2-channel motion vectors are converted to 3-channel RGB values prior to feature embedding. Similar to existing two-stream methods, such as MATNet~\cite{MATNet} and FSNet~\cite{FSNet}, our approach processes the video frame-by-frame. For inferring the segmentation mask of the $i$-th frame, only the $i$-th frame's RGB image and optical flow map are used.

\subsection{Motion-as-Option Network}
To address the vulnerability of existing two-stream architectures to low-quality or confusing optical flow maps, we propose a novel motion-as-option network. This network operates independently of motion cues, reducing its dependency on motion information. The architecture of our proposed motion-as-option network is illustrated in Figure~\ref{figure2}.

\vspace{1mm}
\noindent\textbf{Separate encoders.} Inspired by the complementary advantages of RGB images and optical flow maps, existing two-stream methods use deeply connected encoders to embed each source of information. During the encoding process, appearance features and motion features interact and impose constraints on each other. As these features cannot be embedded independently, the dual encoders function as a single encoder that takes both RGB images and optical flow maps as inputs. Consequently, these methods are inherently reliant on the quality of the optical flow maps, leading to significant errors when low-quality or confusing motion cues are provided.

Unlike the existing two-stream methods, we use two completely separated encoders to embed appearance and motion features independently. In other words, during the feature embedding process, there is no interaction between appearance encoder and motion encoder. Let us denote appearance features as $\{A_k\}_{k=1}^K$ and motion features as $\{M_k\}_{k=1}^K$, where $K$ is the number of blocks in the respective encoders and a higher $k$ value indicates higher-level features. The appearance features are embedded from RGB images, whereas the motion features are embedded from either RGB images or optical flow maps. Then, by simply summing appearance and motion features of each level, fused features $\{X_k\}_{k=1}^K$ can be defined as
\begin{align}
&X_k = A_k + M_k~.
\end{align}

In our encoder, appearance and motion features are embedded independently before being fused. This design enhances the network's robustness against misleading motion cues, as it mitigates potential negative impacts of motion features on appearance features. Moreover, since the motion encoder processes both RGB images and optical flow maps, the network is less prone to overfitting to explicit optical flow maps during training. This also makes the motion-as-option network not always requiring optical flow maps during inference, increasing its overall usability.

\vspace{1mm}
\noindent\textbf{Decoder.} With the fused features, the decoder generates the segmentation mask for the salient object. To convert low-resolution features into high-resolution masks, the fused features are progressively refined through multiple decoding blocks, which are designed similarly to those in TBD~\cite{TBD}. Each decoding block comprises a convolutional layer that integrates features from various embedding levels, a CBAM~\cite{CBAM} layer that enhances feature representations, and an upsampling layer that increases feature resolution. Let $\Psi_k$ represents the $k$-th decoding block, where a higher $k$ value indicates higher resolution. The decoded features $\{D_k\}_{k=1}^K$ are obtained as
\begin{align}
&D_k = 
\begin{cases}
\Psi_k(X_{K-k+1}) &k=1\\ 
\Psi_k(D_{k-1} \oplus X_{K-k+1}) &\text{otherwise}~,
\end{cases}
\end{align}
where $\oplus$ indicates channel concatenation. After passing through $K$ decoding blocks, the final segmentation mask can be obtained after pixel value quantization to 0 or 1.

\begin{algorithm}[t!]
\caption{Feature Fusion for Batch Training}
\begin{algorithmic}[1]
\State \textbf{Input:} $I$, $F$, $V$
\State \textbf{Output:} $A$, $M$, $X$
\State $A_0 \leftarrow I$  
\State $M_0 \leftarrow V * M + (1 - V) * I$  
\For{$k$ in $K$}
\State $A_k \leftarrow$ Embed $A_{k-1}$
\State $M_k \leftarrow$ Embed $M_{k-1}$
\State $X_k \leftarrow A_k + M_k$
\EndFor
\end{algorithmic}
\label{algorithm1}
\end{algorithm}

\begin{algorithm}[t!]
\caption{Adaptive Output Selection}
\begin{algorithmic}[1]
\State \textbf{Input:} $I$, $F$
\State \textbf{Output:} $\omega$
\State Define $\Phi$ as the motion-as-option network
\For{each frame}
\State $\omega_{II} \leftarrow \Phi(I,I)$
\State $\omega_{IF} \leftarrow \Phi(I,F)$
\State Calculate $\alpha$ from $\omega$
\State Compare $\alpha_{II}$ and $\alpha_{IF}$
\State Choose $\omega$ with a higher $\alpha$
\EndFor
\end{algorithmic}
\label{algorithm2}
\end{algorithm}

\subsection{Collaborative Learning Strategy}
\label{cls}
To train the proposed motion-as-option network effectively, we need to prepare training samples both with and without optical flow maps. One straightforward approach would be to randomly discard optical flow maps from VOS training samples. However, to fully utilize the network's capability to function without optical flow maps, we also incorporate SOD samples into the training process.

While leveraging both VOS and SOD samples together is beneficial, it poses a challenge for batch training on GPU devices due to their different formats. To address this issue, we use a simple indexing strategy. First, training samples are randomly selected from both the VOS and SOD datasets according to a pre-defined sampling ratio. Each VOS sample includes an RGB image, an optical flow map, and an object segmentation mask, while each SOD sample contains only an RGB image and an object segmentation mask, requiring careful handling to ensure compatibility during training.

To standardize the formats, we generate void tensors for the SOD samples, which are treated as placeholders for optical flow maps. Furthermore, we introduce a motion validity index for each training sample to indicate the presence of valid motion data. For VOS samples, this index is set to 1 (signifying valid optical flow maps), whereas for SOD samples, it is set to 0 (indicating the absence of motion data). This unified format allows for a consistent feature embedding process, as outlined in Algorithm~\ref{algorithm1}, where $I$ represents the RGB image, $F$ denotes the optical flow map, and $V$ stands for the motion validity index. By adopting this collaborative learning strategy, the motion-as-option network gains more extensive knowledge than it would from training on VOS samples alone.

\begin{figure*}[t]
\centering
\includegraphics[width=1\linewidth]{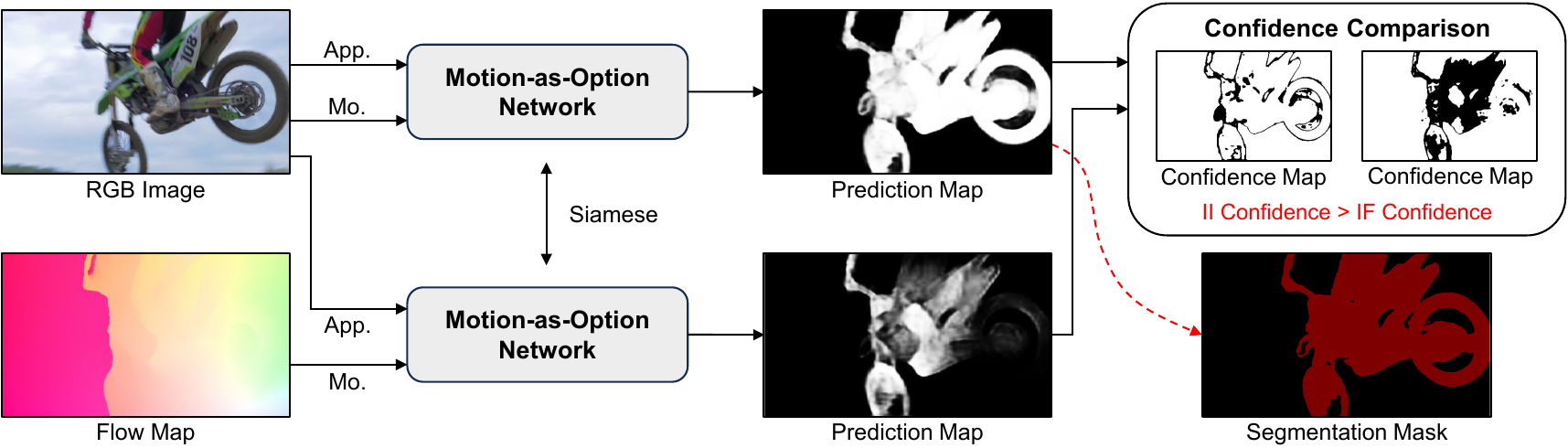}
\vspace{-6mm}
\caption{Visualized pipeline of the adaptive output selection algorithm. The motion-as-option network produces two different outputs: one using an RGB image as the motion input and the other using an optical flow map as the motion input. The final segmentation mask is then obtained by evaluating each output based on the overall confidence scores.}
\label{figure3}
\end{figure*}

\subsection{Adaptive Output Selection}
After training the motion-as-option network using a collaborative learning strategy, the motion encoder gains the ability to process both RGB images and optical flow maps, as it is exposed to both types of data during training. This versatility enables the network to produce two distinct outputs when applied to VOS samples, depending on whether the motion encoder receives an RGB image or an optical flow map, resulting in either image-image (II) inference or image-flow (IF) inference. Preliminary observations~\cite{TMO} reveal that utilizing optical flow maps in the motion encoder does not always produce better results than RGB images. Building on this insight, we introduce a novel adaptive output selection algorithm to identify the optimal source information for the final prediction. The complete pipeline for adaptive output selection is presented in Algorithm~\ref{algorithm2} and visualized in Figure~\ref{figure3}.

Let us denote the output logit map obtained from the motion-as-option network as $\sigma \in \mathbb{R}^{2 \times H \times W}$, where the first and second channels represent background and foreground scores, respectively. By applying a softmax function, we can obtain the foreground prediction map $\omega \in [0,1]^{H \times W}$ as
\begin{align}
\omega^p = \cfrac{\text{exp}(\sigma_{FG}^p)}{\text{exp}(\sigma_{BG}^p) + \text{exp}(\sigma_{FG}^p)},
\end{align}
where $p$ indicates each pixel location. To evaluate the reliability of the prediction results, we calculate the confidence scores of each foreground prediction map using different input domains. Based on a simple clipping function, the confidence map $\phi \in [0,h]^{H \times W}$ is calculated as
\begin{align}
&\phi^p =
\begin{cases}
h - \omega^p &\omega^p < h\\
\omega^p - 1 + h &\omega^p > 1 - h\\
0 &\text{otherwise}~,
\end{cases}
\end{align}
where $h \in [0,0.5]$ is a pre-defined threshold value. Finally, the confidence score $\alpha$ can be obtained as
\begin{align}
\alpha = \lVert\phi\rVert_2~.
\end{align}
The variable $\alpha$ quantifies the overall confidence in the network's predictions by summing the certainty of clearly distinguishable pixels (either definite background or foreground). By comparing $\alpha$ across II and IF inference results, we identify the more reliable output, indicating which input domain is better suited for the motion encoder. The most reliable prediction is then used to generate the final segmentation mask from $\omega$. For example, as shown in Figure~\ref{figure3}, providing an RGB image to the motion encoder results in a higher $\alpha$ value than using an optical flow map. In this case, the final prediction is made without relying on motion cues.

\begin{algorithm}[t!]
\caption{Accelerated Adaptive Output Selection}
\begin{algorithmic}[1]
\State \textbf{Input:} $I$, $F$
\State \textbf{Output:} $\omega$
\State Define $\Phi$ as the motion-as-option network
\For{each frame}
\State A $\leftarrow$ Bind $I$ and $I$ along batch
\State M $\leftarrow$ Bind $I$ and $F$ along batch
\State $\omega \leftarrow \Phi(A,M)$
\State $\omega_{II}$, $\omega_{IF}$ $\leftarrow$ Split $\omega$ along batch
\State Calculate $\alpha$ from $\omega$
\State Compare $\alpha_{II}$ and $\alpha_{IF}$
\State Choose $\omega$ with a higher $\alpha$
\EndFor
\end{algorithmic}
\label{algorithm3}
\end{algorithm}

The proposed adaptive output selection mechanism significantly enhances system stability by enabling the final segmentation mask to be selected from two independently inferred outputs (II inference and IF inference) based on a confidence score comparison. However, this approach introduces additional computational overhead during inference, which inevitably slows down inference speed and impacts the system's applicability. To address this limitation, we propose an acceleration protocol leveraging batch computation on GPU devices. Instead of executing the system twice, we bind the source data along the batch dimension before the feature embedding process. Subsequently, all encoding and decoding steps are executed in a single pass using the batch protocol. Once the final prediction map $\omega$ is obtained, the results are split into individual elements corresponding to each batch entry. The pipeline for the accelerated adaptive output selection is detailed in Algorithm~\ref{algorithm3}.

\vspace{-1mm}
\subsection{Implementation Details}
\noindent\textbf{Optical flow map.} To incorporate motion cues alongside appearance cues, we utilize optical flow maps as source information. To generate the flow map for a given frame $i$, we treat frame $i$ as the starting frame and frame $i+1$ as the target frame. If $i$ is the last frame in the video, frame $i-1$ is used as the target frame instead. For optical flow estimation, we employ RAFT~\cite{RAFT} pre-trained on the Sintel~\cite{Sintel} dataset. The flow maps are generated while preserving the original resolution of the data samples. To avoid redundant processing during training and testing, these optical flow maps are pre-generated and stored, rather than being computed dynamically during system execution.

\vspace{1mm}
\noindent\textbf{Encoder.} We adopt ResNet-101~\cite{resnet} and MiT-b1~\cite{mit} as our backbone encoders. As pixel-level prediction is required, only the first four blocks are detached and used for feature embedding. The features extracted from the $k$-th block have the scale of $1/2^{k+1}$ compared to the input resolution, where total block number $K$ is 4. The encoders are initialized with the weights learned from ImageNet~\cite{imagenet}.

\vspace{1mm}
\noindent\textbf{Confidence map.} To evaluate the certainty level of predictions, confidence maps are generated from the foreground prediction maps. During this process, pixels that are not clearly distinguishable are first filtered out before calculating the confidence score. The threshold value $h$ is empirically set to 0.1; however, no significant performance difference is observed with different $h$ values.

\begin{table*}[t]
\centering 
\small
\caption{Comparison of training data details among state-of-the-art methods.}
\vspace{-2mm}
\begin{tabular}{lP{1.7cm}P{9.5cm}P{1.2cm}P{1.2cm}}
\toprule
Method &Publication &Training Data Description &Video \# &Image \#\\
\midrule
MATNet~\cite{MATNet} &AAAI'20 &YouTube-VOS 2018~\cite{YTVOS} + DAVIS 2016~\cite{DAVIS} &3,501 &0\\
RTNet~\cite{RTNet} &CVPR'21 &DUTS~\cite{DUTS} + DAVIS 2016~\cite{DAVIS} &30 &15,572\\
FSNet~\cite{FSNet} &ICCV'21 &DUTS~\cite{DUTS} + DAVIS 2016~\cite{DAVIS} + FBMS~\cite{FBMS} &59 &15,572\\
D$^2$Conv3D~\cite{D^2Conv3D} &WACV'22 &Kinetics400~\cite{kinetics} + Sports-1M~\cite{sports} + DAVIS 2017~\cite{DAVIS17} &1.3M &0\\
IMP~\cite{IMP} &AAAI'22 &DUTS~\cite{DUTS} (excluding PFPN~\cite{PFPN} and STM~\cite{STM}) &0 &15,572\\
HFAN~\cite{HFAN} &ECCV'22 &YouTube-VOS 2018~\cite{YTVOS} + DAVIS 2016~\cite{DAVIS} &3,501 &0\\
PMN~\cite{PMN} &WACV'23 &DUTS~\cite{DUTS} + DAVIS 2016~\cite{DAVIS} &30 &15,572\\
OAST~\cite{OAST} &ICCV'23 &YouTube-VOS 2018~\cite{YTVOS} + DAVIS 2016~\cite{DAVIS} + YouTube-Objects~\cite{YTOBJ} &3,627 &0\\
SimulFlow~\cite{simulflow} &ACMMM'23 &DUTS~\cite{DUTS} + YouTube-VOS 2018~\cite{YTVOS} + DAVIS 2016~\cite{DAVIS} &3,501 &15,572\\
GFA~\cite{GFA} &AAAI'24 &YouTube-VOS 2018~\cite{YTVOS} + DAVIS 2016~\cite{DAVIS} &3,501 &0\\
DATTT~\cite{DATTT} &CVPR'24 &YouTube-VOS 2018~\cite{YTVOS} + Test-Time Video &3,471+1 &0\\
DPA~\cite{DPA} &CVPR'24 &DUTS~\cite{DUTS} + YouTube-VOS 2018~\cite{YTVOS} + DAVIS 2016~\cite{DAVIS} &3,501 &15,572\\
GSA-Net~\cite{GSA-Net} &CVPR'24 &DUTS~\cite{DUTS} + YouTube-VOS 2018~\cite{YTVOS} + DAVIS 2016~\cite{DAVIS} &3,501 &15,572\\
\midrule
\textbf{TMO}, \textbf{TMO++} & &DUTS~\cite{DUTS} + DAVIS 2016~\cite{DAVIS} &30 &15,572\\
\bottomrule
\end{tabular}
\label{table2}
\end{table*}

\subsection{Network Training}
\label{training}
\noindent\textbf{Data preparation.} To enhance training data diversity and fully exploit the unique properties of the motion-as-option network, we incorporate both VOS and SOD samples during network training, as described in Section~\ref{cls}. For VOS samples, we exclusively use the DAVIS 2016~\cite{DAVIS} training set, adhering to the common protocol in unsupervised VOS, which excludes the FBMS~\cite{FBMS} training set. For SOD samples, we use both the training and testing sets from DUTS~\cite{DUTS}. Our training data is randomly sampled from VOS and SOD datasets with fixed probabilities of 25\% and 75\%, respectively.

\vspace{1mm}
\noindent\textbf{Training details.} As the training objective, we adopt the conventional cross-entropy loss function. Additionally, we use the Adam optimizer with a learning rate of 1e-5 without learning rate decay. The batch size is set to 16. Following recent VOS algorithms, we freeze all batch normalization layers during network training. For all model versions, network training is implemented on two GeForce RTX 2080 Ti GPUs and takes less than 20 hours.

\section{Experiments}
In Section~\ref{dataset}, we describe the datasets used in this study. Evaluation metrics and comparisons of training data are provided in Sections~\ref{metric} and~\ref{data}, respectively. Sections~\ref{qualitative} and~\ref{quantitative} present qualitative and quantitative comparisons of our proposed approach with state-of-the-art methods. The effectiveness of each proposed component is extensively analyzed in Section~\ref{analysis}. Our baseline method is referred to as TMO, while its enhanced version with the adaptive output selection algorithm is denoted as TMO++.

\subsection{Datasets}
\label{dataset}
To confirm the validity of our proposed approach, we use five datasets that are widely adopted for unsupervised VOS. For network training, DUTS~\cite{DUTS} and DAVIS 2016~\cite{DAVIS} are utilized. For network testing, DAVIS 2016, FBMS~\cite{FBMS}, YouTube-Objects~\cite{YTOBJ}, and Long-Videos~\cite{LVID} are used.

\subsection{Evaluation Metrics}
\label{metric}
To measure the segmentation performance of unsupervised VOS methods, we employ evaluation protocols commonly used in general image segmentation tasks. Two widely adopted metrics are $\mathcal{J}$ and $\mathcal{F}$. The $\mathcal{J}$ metric, also known as intersection-over-union (IoU), measures the region similarity and is defined as
\begin{align}
&\mathcal{J} = \left| \frac{M_{gt} \cap M_{pred}}{M_{gt} \cup M_{pred}} \right|~,
\label{eq3}
\end{align}
where $M_{gt}$ and $M_{pred}$ denote the ground truth and predicted segmentation masks, respectively. $\mathcal{F}$ metric is similar to $\mathcal{J}$ metric, but is calculated specifically for object boundaries. It is computed as
\begin{align}
&\text{Precision} = \left| \frac{M_{gt} \cap M_{pred}}{M_{pred}} \right|~,
\end{align}
\begin{align}
&\text{Recall} = \left| \frac{M_{gt} \cap M_{pred}}{M_{gt}} \right|~,
\end{align}
\begin{align}
&\mathcal{F} = \frac{2 \times \text{Precision} \times \text{Recall}}{\text{Precision} + \text{Recall}}~.
\end{align}
Additionally, the $\mathcal{G}$ metric, which is the average of $\mathcal{J}$ and $\mathcal{F}$, is also widely used to evaluate VOS performance.

\subsection{Training Data Comparison}
\label{data}
To enable a fair comparison among different methods, we first outline the detailed training data settings of both existing and proposed approaches in Table~\ref{table2}. The ``Video \#" refers to the count of video sequences utilized for network training, while the ``Image \#" indicates the number of image samples employed. As shown in the table, our approach demonstrates a significant advantage by requiring a relatively small amount of video data compared to existing methods.

\begin{figure*}[t]
\centering
\includegraphics[width=1\linewidth]{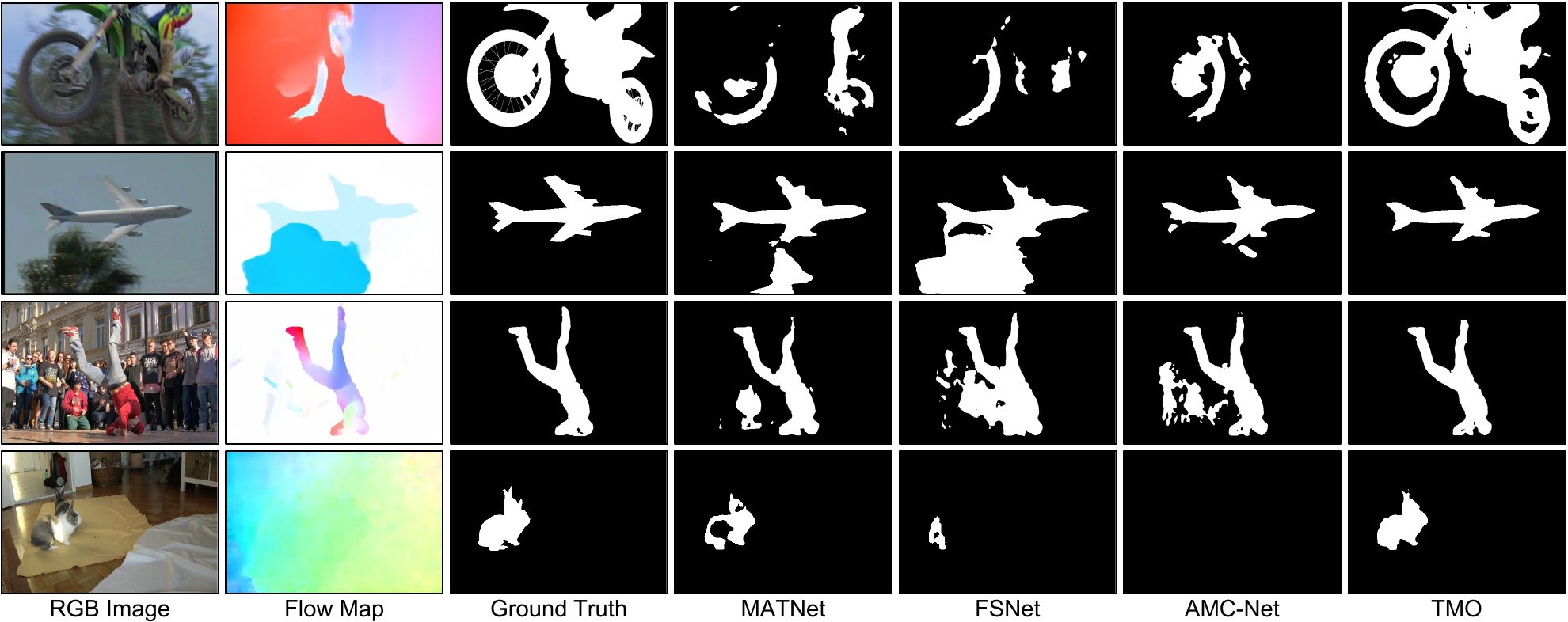}
\vspace{-6mm}
\caption{Qualitative comparison of flow-based approaches in challenging flow scenarios.}
\label{figure4}
\end{figure*}

\begin{figure*}[t]
\centering
\includegraphics[width=1\linewidth]{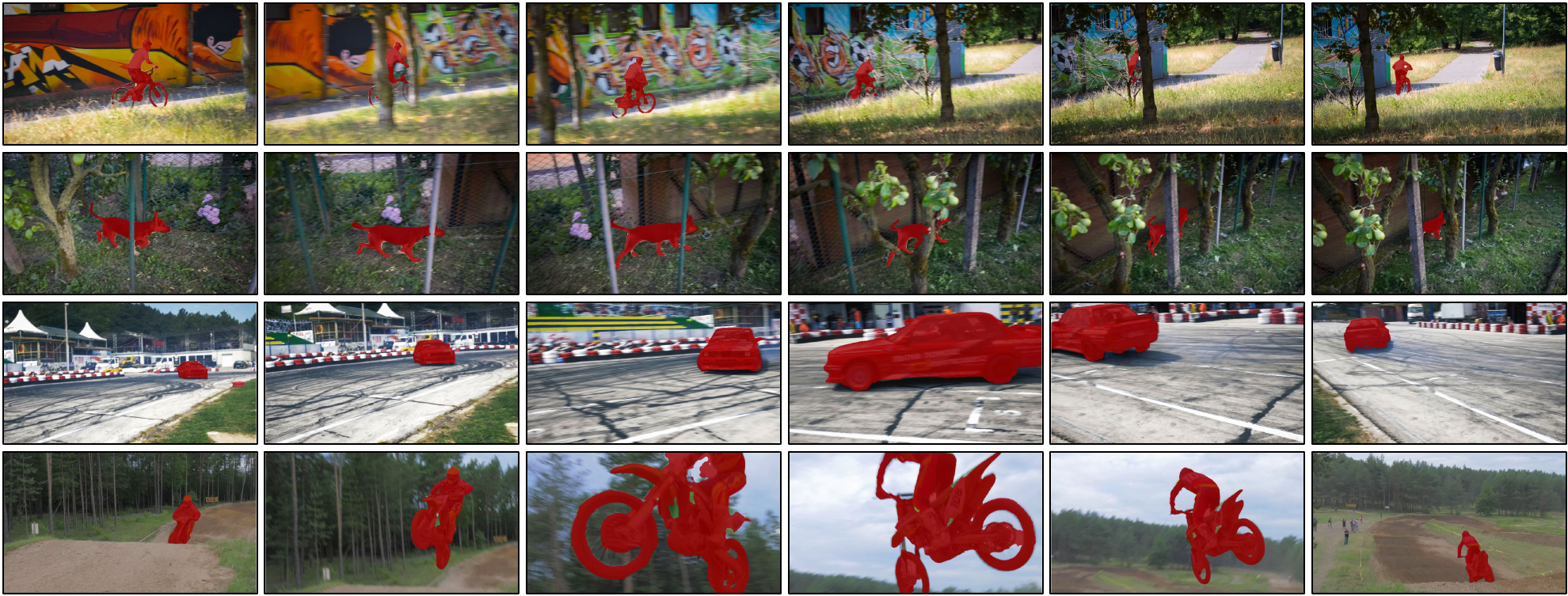}
\vspace{-6mm}
\caption{Automatic object segmentation results on videos.}
\label{figure5}
\end{figure*}

\subsection{Qualitative Results}
\label{qualitative}
In Figure~\ref{figure4}, we compare our method with state-of-the-art approaches, including MATNet~\cite{MATNet}, FSNet~\cite{FSNet}, and AMC-Net~\cite{AMC-Net}. To ensure fairness, we select methods that employ a two-stream architecture and per-frame processing. The top three examples highlight scenarios where optical flow maps exhibit distracting properties due to structural inconsistencies with the target object. The first example shows fast motion, causing optical flow maps to inadequately capture the object’s properties. The second involves occlusion by other objects, generating disruptive motion. The third shows significant background motion, introducing distractions in the motion cues. The last example depicts cases where optical flow maps provide little useful information due to minimal object motion. In all these cases, the optical flow maps fail to deliver reliable cues for the network, emphasizing the robustness of our method against low-quality motion inputs. Our approach demonstrates significantly better stability and prediction quality in such conditions.

Furthermore, Figure~\ref{figure5} visualizes video-level results of our method. The top two scenarios involve severe object occlusions, while the remaining two depict fast-moving target objects. In both types of challenging scenarios, our approach consistently generates accurate and stable masks across video frames. This demonstrates its effectiveness in handling diverse situations, where occlusions and rapid motion typically complicate segmentation tasks.

\subsection{Quantitative Results}
\label{quantitative}
To assess the performance of our proposed approach, we compare it with state-of-the-art methods across four benchmark datasets: the DAVIS 2016~\cite{DAVIS} validation set, the FBMS~\cite{FBMS} test set, the YouTube-Objects~\cite{YTOBJ} dataset, and the Long-Videos~\cite{LVID} dataset. For a fair comparison, we also report results with test-time augmentation. Following the strategy employed in HFAN~\cite{HFAN}, we utilize a multi-scale and flipping ensemble, combining three resolutions (288p, 384p, and 672p) with both horizontal and vertical flipping, resulting in 12 distinct inference steps. All evaluations are performed on a single GeForce RTX 2080 Ti GPU.

\begin{table*}
\centering 
\caption{Quantitative evaluation on the DAVIS 2016 validation set and FBMS test set. OF and PP indicate the use of optical flow estimation model and post-processing technique, respectively. $\ast$ denotes speed calculated on our hardware. $\dagger$ indicates the use of test-time augmentation strategy.}
\vspace{-2mm}
\small
\begin{tabular}{p{2.2cm}P{1.9cm}cP{1.5cm}P{0.5cm}P{0.5cm}P{0.7cm}P{0.7cm}P{0.7cm}P{0.7cm}P{0.7cm}}
\toprule
\multicolumn{7}{c}{} &\multicolumn{3}{c}{DAVIS 2016} &\multicolumn{1}{c}{FBMS}\\
\cmidrule(lr){8-11}
Method &Publication &Backbone &Resolution &OF &PP &fps &$\mathcal{G}_\mathcal{M}$ &$\mathcal{J}_\mathcal{M}$ &$\mathcal{F}_\mathcal{M}$ &$\mathcal{J}_\mathcal{M}$\\
\midrule
MATNet~\cite{MATNet} &AAAI'20 &ResNet-101~\cite{resnet} &473$\times$473 &\checkmark &\checkmark &20.0 &81.6 &82.4 &80.7 &76.1\\
RTNet~\cite{RTNet} &CVPR'21 &ResNet-101~\cite{resnet} &384$\times$672 &\checkmark &\checkmark &- &85.2 &85.6 &84.7 &-\\
FSNet~\cite{FSNet} &ICCV'21 &ResNet-50~\cite{resnet} &352$\times$352 &\checkmark &\checkmark &12.5 &83.3 &83.4 &83.1 &-\\
TransportNet~\cite{TransportNet} &ICCV'21 &ResNet-101~\cite{resnet} &512$\times$512 &\checkmark & &12.5 &84.8 &84.5 &85.0 &78.7\\
AMC-Net~\cite{AMC-Net} &ICCV'21 &ResNet-101~\cite{resnet} &384$\times$384 &\checkmark &\checkmark &17.5 &84.6 &84.5 &84.6 &76.5\\
D$^2$Conv3D~\cite{D^2Conv3D} &WACV'22 &ir-CSN-152~\cite{csn} &480$\times$854 & & &- &86.0 &85.5 &86.5 &-\\
IMP~\cite{IMP} &AAAI'22 &ResNet-50~\cite{resnet} &- & & &1.79 &85.6 &84.5 &86.7 &77.5\\
HFAN~\cite{HFAN} &ECCV'22 &MiT-b1~\cite{mit} &512$\times$512 &\checkmark & &18.4$^\ast$ &86.7 &86.2 &87.1 &-\\
FEM-Net~\cite{vos5} &TCSVT'22 &ResNet-101~\cite{resnet} &473$\times$473 &\checkmark & &7.00 &79.9 &76.9 &78.4 &78.5\\
IMCNet~\cite{vos6} &TCSVT'22 &ResNet-101~\cite{resnet} &480$\times$480 & & &20.0 &81.9 &82.7 &81.1 &-\\
PMN~\cite{PMN} &WACV'23 &VGG-16~\cite{vgg} &352$\times$352 &\checkmark & &41.3$^\ast$ &85.9 &85.4 &86.4 &77.7\\
OAST~\cite{OAST} &ICCV'23 &3D-ResNet-101~\cite{3dresnet} &384$\times$640 &\checkmark & &- &85.9 &85.4 &86.3 &81.7\\
SimulFlow~\cite{simulflow} &ACMMM'23 &MiT-b1~\cite{mit} &384$\times$384 &\checkmark & &53.2 &86.7 &86.3 &87.4 &80.1\\
HCPN~\cite{HCPN} &TIP'23 &ResNet-101~\cite{resnet} &512$\times$512 &\checkmark &\checkmark &2.00 &85.6 &85.8 &85.4 &78.3\\
GFA~\cite{GFA} &AAAI'24 &- &512$\times$512 &\checkmark & &- &\textbf{88.2} &\underline{87.4} &\underline{88.9} &82.4\\
DATTT~\cite{DATTT} &CVPR'24 &MiT-b1~\cite{mit} &512$\times$512 &\checkmark & &- &87.0 &86.0 &87.9 &74.9\\
DPA~\cite{DPA} &CVPR'24 &VGG-16~\cite{vgg} &512$\times$512 &\checkmark & &19.5$^\ast$ &\underline{87.6} &86.8 &88.4 &\textbf{83.4}\\
GSA-Net~\cite{GSA-Net} &CVPR'24 &MiT-b2~\cite{mit} &512$\times$512 &\checkmark & &31.0$^\ast$ &\textbf{88.2} &\underline{87.4} &\textbf{89.0} &82.3\\
\midrule
\textbf{TMO} & &ResNet-101~\cite{resnet} &384$\times$384 &\checkmark & &43.2$^\ast$ &86.1 &85.6 &86.6 &79.9\\
\textbf{TMO++} & &ResNet-101~\cite{resnet} &384$\times$384 &\checkmark & &36.5$^\ast$ &86.1 &85.6 &86.6 &81.4\\
\textbf{TMO++}$^\dagger$ & &ResNet-101~\cite{resnet} &Multiple &\checkmark & &2.3$^\ast$ &87.2 &\underline{87.4} &87.0 &81.7\\
\midrule
\textbf{TMO} & &MiT-b1~\cite{mit} &384$\times$384 &\checkmark & &\textbf{66.7}$^\ast$ &87.2 &86.5 &87.8 &80.0\\
\textbf{TMO++} & &MiT-b1~\cite{mit} &384$\times$384 &\checkmark & &\underline{54.3}$^\ast$ &87.2 &86.5 &87.8 &\underline{83.2}\\
\textbf{TMO++}$^\dagger$ & &MiT-b1~\cite{mit} &Multiple &\checkmark & &4.3$^\ast$ &\textbf{88.2} &\textbf{88.0} &88.3 &\textbf{83.4}\\
\bottomrule
\end{tabular}
\label{table3}
\end{table*}

\begin{table*}[t]
\centering 
\small
\caption{Quantitative evaluation on the YouTube-Objects dataset. Performance is reported with the $\mathcal{J}$ mean. $\dagger$ indicates the use of test-time augmentation strategy.}
\vspace{-2mm}
\begin{tabular}{lcP{7.5mm}P{7.5mm}P{7.5mm}P{7.5mm}P{7.5mm}P{7.5mm}P{7.5mm}P{7.5mm}P{8mm}P{7.5mm}|P{7.5mm}}
\toprule
Method &Backbone &Aero. &Bird &Boat &Car &Cat &Cow &Dog &Horse &Motor. &Train &Mean\\
\midrule
AGS~\cite{AGS} &ResNet-101~\cite{resnet} &\textbf{87.7} &76.7 &\textbf{72.2} &78.6 &69.2 &64.6 &73.3 &64.4 &62.1 &48.2 &69.7\\
COSNet~\cite{COSNet} &DeepLabv3~\cite{deeplabv3} &81.1 &75.7 &71.3 &77.6 &66.5 &69.8 &76.8 &\underline{67.4} &\underline{67.7} &46.8 &70.5\\
AGNN~\cite{AGNN} &DeepLabv3~\cite{deeplabv3} &71.1 &75.9 &70.7 &78.1 &67.9 &69.7 &77.4 &67.3 &\textbf{68.3} &47.8 &70.8\\
MATNet~\cite{MATNet} &ResNet-101~\cite{resnet} &72.9 &77.5 &66.9 &79.0 &73.7 &67.4 &75.9 &63.2 &62.6 &51.0 &69.0\\
RTNet~\cite{RTNet} &ResNet-101~\cite{resnet} &84.1 &80.2 &70.1 &79.5 &71.8 &70.1 &71.3 &65.1 &64.6 &53.3 &71.0\\
AMC-Net~\cite{AMC-Net} &ResNet-101~\cite{resnet} &78.9 &80.9 &67.4 &\underline{82.0} &69.0 &69.6 &75.8 &63.0 &63.4 &57.8 &71.1\\
HFAN$^\dagger$~\cite{HFAN} &MiT-b1~\cite{mit} &84.7 &80.0 &\underline{72.0} &76.1 &76.0 &71.2 &76.9 &\textbf{71.0} &64.3 &\textbf{61.4} &73.4\\
IMCNet~\cite{vos5} &ResNet-101~\cite{resnet} &81.2 &78.3 &67.9 &78.5 &70.4 &67.1 &73.1 &63.0 &63.3 &56.8 &70.0\\
HCPN~\cite{HCPN} &ResNet-101~\cite{resnet} &84.5 &79.6 &67.3 &\textbf{87.8} &74.1 &71.2 &76.5 &66.2 &65.8 &\underline{59.7} &73.3\\
\midrule
\textbf{TMO} &ResNet-101~\cite{resnet} &85.7 &80.0 &70.1 &78.0 &73.6 &70.3 &76.8 &66.2 &58.6 &47.0 &71.5\\
\textbf{TMO++} &ResNet-101~\cite{resnet} &84.2 &80.2 &71.9 &78.6 &74.8 &\underline{72.3} &79.2 &66.9 &63.3 &47.2 &73.1\\
\textbf{TMO++}$^\dagger$ &ResNet-101~\cite{resnet} &\underline{85.8} &80.4 &71.6 &78.8 &76.5 &\textbf{72.9} &\underline{79.9} &66.6 &64.4 &47.7 &\textbf{73.7}\\
\midrule
\textbf{TMO} &MiT-b1~\cite{mit} &81.8 &\underline{84.5} &68.0 &77.9 &74.7 &67.9 &78.0 &65.0 &56.2 &49.6 &71.1\\
\textbf{TMO++} &MiT-b1~\cite{mit} &83.2 &\textbf{85.0} &69.9 &78.1 &\underline{77.3} &70.6 &\textbf{80.4} &64.9 &61.1 &50.8 &73.1\\
\textbf{TMO++}$^\dagger$ &MiT-b1~\cite{mit} &83.2 &\textbf{85.0} &71.2 &77.5 &\textbf{78.0} &71.3 &\textbf{80.4} &65.3 &62.3 &50.6 &\underline{73.6}\\
\bottomrule
\end{tabular}
\label{table4}
\end{table*}

\begin{table}[t!]
\centering 
\small
\caption{Quantitative evaluation on the Long-Videos dataset. SS and US denote semi-supervised and unsupervised, respectively. $\dagger$ indicates the use of test-time augmentation strategy.}
\vspace{-2mm}
\begin{tabular}{p{2cm}cP{0.8cm}P{0.8cm}}
\toprule
Method &Backbone &Type &$\mathcal{J}_\mathcal{M}$\\
\midrule
A-GAME~\cite{A-GAME} &ResNet-101~\cite{resnet} &SS &50.0\\
STM~\cite{STM} &ResNet-50~\cite{resnet} &SS &\underline{79.1}\\
CFBI~\cite{CFBI} &DeepLabv3+~\cite{deeplabv3p} &SS &50.9\\
AFB-URR~\cite{LVID} &ResNet-50~\cite{resnet} &SS &\textbf{82.7}\\
\midrule
AGNN~\cite{AGNN} &DeepLabv3~\cite{deeplabv3} &US &68.3\\
MATNet~\cite{MATNet} &ResNet-101~\cite{resnet} &US &66.4\\
3DC-Seg~\cite{3DC-Seg} &ir-CSN-152~\cite{csn} &US &34.2\\
HFAN$^\dagger$~\cite{HFAN} &MiT-b1~\cite{mit} &US &\underline{74.9}\\
\midrule
\textbf{TMO} &MiT-b1~\cite{mit} &US &71.8\\
\textbf{TMO++} &MiT-b1~\cite{mit} &US &74.8\\
\textbf{TMO++}$^\dagger$ &MiT-b1~\cite{mit} &US &\textbf{75.8}\\
\bottomrule
\end{tabular}
\label{table5}
\end{table}

\vspace{1mm} 
\noindent\textbf{DAVIS 2016.} 
Table~\ref{table3} summarizes the performance on the DAVIS 2016 validation set. For a fair comparison, we explicitly mark methods that incorporate external post-processing techniques, such as fully-connected dense CRF~\cite{densecrf} or instance pruning~\cite{AD-Net}. Where available, inference speed is also reported to evaluate system efficiency, although the reported speeds do not account for the time spent on optical flow generation or post-processing.

D$^2$Conv3D~\cite{D^2Conv3D} and IMP~\cite{IMP} demonstrate strong performance with $\mathcal{G}$ scores of 86.0\% and 85.6\%, respectively, using only RGB images. However, their reliance on processing multiple video frames limits their applicability in online scenarios. Among methods with comparable backbones to ours, DATTT~\cite{DATTT} achieves the highest performance. Despite not employing post-processing techniques, our method achieves superior prediction accuracy and inference speed while maintaining online availability. Notably, TMO with MiT-b1~\cite{mit} achieves a $\mathcal{G}$ score of 87.2\% with exceptional inference speed, showcasing its efficiency and effectiveness. When using an ensemble technique, performance is further improved, reaching a $\mathcal{G}$ score of 88.2\%.

\vspace{1mm} 
\noindent\textbf{FBMS.} 
Table~\ref{table3} also reports the quantitative evaluation on the FBMS test set. A unique feature of the FBMS dataset is that video sequences often contain multiple salient objects that must be detected simultaneously. With adaptive output selection, our method achieves a high $\mathcal{J}$ score of 83.4\%, demonstrating strong performance in this challenging scenario.

\vspace{1mm} 
\noindent\textbf{YouTube-Objects.} 
Table~\ref{table4} presents a quantitative comparison of various VOS approaches on the YouTube-Objects dataset. Model performance is evaluated using both per-class accuracy (indicating the mean score across sequences in each class) and overall accuracy (averaging scores across all sequences). Among existing methods, HFAN with the MiT-b1 backbone achieves an overall $\mathcal{J}$ score of 73.4\% with test-time augmentation. Using the same backbone and test-time augmentation strategy, TMO++ surpasses HFAN with an overall $\mathcal{J}$ score of 73.6\%.

\vspace{1mm}
\noindent\textbf{Long-Videos.} 
Finally, Table~\ref{table5} shows the performance on the Long-Videos dataset. Among unsupervised VOS methods, TMO++ with test-time augmentation achieves the best performance with a $\mathcal{J}$ score of 75.8\%, highlighting its superiority in real-world scenarios. Interestingly, our method outperforms some semi-supervised VOS methods, which typically have an advantage over unsupervised methods.

\subsection{Analysis} 
\label{analysis} 
In this section, we analyze the network training and testing protocols. We use the $\mathcal{G}$ score for the DAVIS 2016~\cite{DAVIS} validation set, and the $\mathcal{J}$ score for the FBMS~\cite{FBMS} test set and YouTube-Objects~\cite{YTOBJ} dataset. ResNet-101~\cite{resnet} is used as the backbone encoder.

\vspace{1mm} 
\noindent\textbf{Collaborative learning strategy.} 
We validate our collaborative learning strategy through training data construction, as shown in Table~\ref{table6}. In model \Romannum{1}, where only VOS samples are used for training, the network tends to overfit, resulting in reduced performance. Pre-training on SOD samples and fine-tuning on VOS samples (model \Romannum{2}) improves performance but still falls short. This occurs because the network forgets knowledge from SOD samples. In contrast, our collaborative learning strategy in model \Romannum{3} achieves the highest performance across all datasets, with significant improvements on FBMS and YouTube-Objects, where flow maps are less reliable than in DAVIS. By incorporating both VOS and SOD samples during training, the network learns high-quality details from VOS while benefiting from the broad insights of SOD samples.

Additionally, in Table~\ref{table7}, we evaluate the impact of varying the ratio of VOS and SOD samples. The results show that combining VOS and SOD samples improves performance compared to using either type alone. The optimal ratio is 25\% VOS to 75\% SOD, with minimal performance differences across other ratios.

\begin{table}[t!]
\centering 
\caption{Ablation study on training and testing protocols. Training indicates which training protocols are adopted for network training. Testing indicates input of the motion encoder during inference. D, F, and Y denote DAVIS 2016 validation set, FBMS test set, and YouTube-Objects dataset, respectively.}
\vspace{-2mm}
\small
\begin{tabular}{c|c|c|P{0.6cm}P{0.6cm}P{0.6cm}}
\toprule
Version &Training &Testing &D &F &Y\\
\midrule
\Romannum{1} &VOS &Flow &76.5 &59.2 &57.2\\
\Romannum{2} &SOD $\rightarrow$ VOS &Flow &82.1 &74.7 &63.0\\
\Romannum{3} &VOS~\&~SOD &Flow &86.1 &79.9 &71.5\\
\midrule
\Romannum{4} &VOS &Image &63.7 &52.8 &52.6\\
\Romannum{5} &SOD $\rightarrow$ VOS &Image &73.0 &76.8 &69.3\\
\Romannum{6} &VOS~\&~SOD &Image &80.0 &80.0 &73.1\\
\bottomrule
\end{tabular}
\label{table6}
\end{table}

\begin{table}[t!]
\centering 
\caption{Ablation study on the dataset mixing ratio. D, F, and Y denote DAVIS 2016 validation set, FBMS test set, and YouTube-Objects dataset, respectively.}
\vspace{-2mm}
\small
\begin{tabular}{c|P{1cm}|P{1cm}|P{0.6cm}P{0.6cm}P{0.6cm}}
\toprule
Version &VOS &SOD &D &F &Y\\
\midrule
\Romannum{1} &0\% &100\% &82.8 &78.4 &71.0\\
\Romannum{2} &25\% &75\% &86.1 &79.9 &71.5\\
\Romannum{3} &50\% &50\% &85.8 &80.0 &71.4\\
\Romannum{4} &75\% &25\% &85.8 &79.5 &70.9\\
\Romannum{5} &100\% &0\% &76.5 &59.2 &57.2\\
\bottomrule
\end{tabular}
\label{table7}
\end{table}

\begin{figure*}[t]
\centering
\includegraphics[width=1\linewidth]{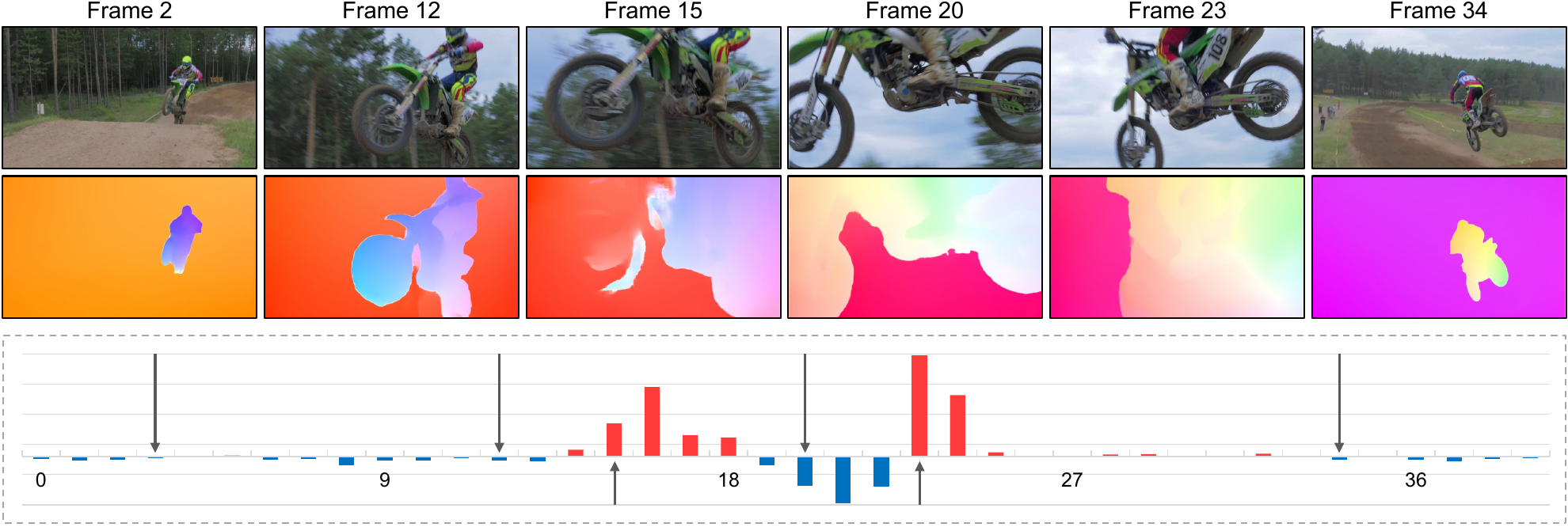}
\vspace{-6mm}
\caption{Visualized confidence scores of a sample video sequence. The $x$-axis represents the frame number, while the $y$-axis shows the difference in confidence scores between using RGB images as motion input and using optical flow maps as motion input. Positive difference values indicate that using RGB images results in higher confidence scores compared to using optical flow maps, and vice versa.}
\label{figure6}
\end{figure*}

\vspace{1mm}
\noindent\textbf{Motion dependency.}
We report the performance of models when RGB images are used as motion encoder input during testing. As shown in models \Romannum{4}, \Romannum{5}, and \Romannum{6} in Table~\ref{table6}, performance decreases when RGB images replace optical flow maps in the DAVIS 2016 validation set. Model \Romannum{6} shows the smallest decrement, indicating low dependency on motion cues during inference. Interestingly, on FBMS and YouTube-Objects, motion cues are less effective than on DAVIS, and omitting motion cues can improve performance. Specifically, model \Romannum{6} achieves 0.1\% and 1.6\% higher scores on FBMS and YouTube-Objects, respectively.

\begin{table}[t!]
\centering 
\small
\caption{Ablation study on test-time output fusion method. D, F, and Y denote DAVIS 2016 validation set, FBMS test set, and YouTube-Objects dataset, respectively.}
\vspace{-2mm}
\begin{tabular}{c|c|c|ccc}
\toprule
Version &Abbr. &Description &D &F &Y\\
\midrule
\Romannum{1} &TMO &IF Inference &86.1 &79.9 &71.5\\
\Romannum{2} &- &II Inference &80.0 &80.0 &73.1\\
\midrule
\Romannum{3} &- &Input Ensemble &83.7 &80.3 &72.8\\
\Romannum{4} &- &Feature Ensemble &84.6 &80.9 &73.3\\
\Romannum{5} &- &Output Ensemble &84.7 &81.0 &73.2\\
\midrule
\Romannum{6} &TMO++ &Output Selection &86.1 &81.4 &73.1\\
\bottomrule
\end{tabular}
\label{table8}
\end{table}

\begin{table}[t!]
\centering 
\caption{Quantitative analysis on adaptive output selection. Image and flow indicate the selection ratios of RGB images and optical flow maps as motion encoder input, respectively.}
\vspace{-2mm}
\small
\begin{tabular}{c|P{1.5cm}P{1.5cm}}
\toprule
Dataset &Image (\%) &Flow (\%)\\
\midrule
DAVIS 2016~\cite{DAVIS} &3.78 &96.2\\
FBMS~\cite{FBMS} &37.6 &62.4\\
YouTube-Objects~\cite{YTOBJ} &38.8 &61.2\\
\bottomrule
\end{tabular}
\vspace{-1mm}
\label{table9}
\end{table}

\vspace{1mm} 
\noindent\textbf{Adaptive output selection.} 
As shown in Table~\ref{table6}, motion cues do not always enhance performance, highlighting that the optimal input for the motion encoder varies by context. II inference can outperform IF inference in certain scenarios. When optical flow maps are high quality, they are superior for the motion encoder; however, when their quality is low, RGB images are more effective. To address this variability, we propose an adaptive output selection algorithm that dynamically determines the optimal source.

Figure~\ref{figure6} illustrates per-frame confidence score differences. Results show that even within a single video, the optimal input fluctuates. For instance, at frames 15 and 23, RGB images yield higher confidence scores than flow maps, while at frame 20, optical flow maps provide higher confidence. This qualitative analysis aligns with quantitative findings, supporting the effectiveness of the adaptive output selection.

Table~\ref{table8} compares different output processing methods, including models that rely on a single source (model \Romannum{1} and model \Romannum{2}) or fuse both RGB images and flow maps before, during, or after decoding (models \Romannum{3}, \Romannum{4}, and \Romannum{5}). While fusion methods show some improvements on FBMS and YouTube-Objects, performance on the DAVIS 2016 validation set is significantly lower than model \Romannum{1}. The performance gap highlights the negative influence of ensemble method, whereas the adaptive output selection avoids performance degradation by using only the optimal output for final predictions.

Table~\ref{table9} shows the selection ratios of motion encoder inputs across datasets. In the DAVIS 2016 validation set, RGB images are chosen for only 3.78\% of frames, while they are selected for 37.6\% and 38.8\% of frames in the FBMS and YouTube-Objects datasets, respectively. These results suggest that flow maps are particularly effective for DAVIS but less so for the other datasets.

\begin{table}[t!]
\centering 
\small
\caption{Detailed analysis of the acceleration for adaptive output selection. B denotes the batch size.}
\vspace{-2mm}
\begin{tabular}{P{2.4cm}|P{1.18cm}P{1.18cm}P{1.18cm}|P{0.6cm}}
\toprule
Description &Enc$_{app}$ \# &Enc$_{mo}$ \# &Dec \# &fps\\
\midrule
IF or II Inference &1 (B=1) &1 (B=1) &1 (B=1) &43.2\\
\midrule
Output Selection &1 (B=1) &2 (B=1) &2 (B=1) &26.3\\
+ Acceleration &1 (B=1) &1 (B=2) &1 (B=2) &36.5\\
\bottomrule
\end{tabular}
\label{table10}
\end{table}

\vspace{1mm} 
\noindent\textbf{Accelerating adaptive output selection.} 
The adaptive output selection improves stability but reduces efficiency due to increased computational costs. Each inference requires two passes through the motion encoder and decoding stages, which creates a bottleneck. To address this, we implement a batch acceleration strategy, processing both IF and II inferences in a single pass on GPU devices. This reduces the slowdown from 39.1\% to 15.5\%, improving efficiency while maintaining accuracy, as shown in Table~\ref{table10}.

\begin{table}[t!]
\centering 
\caption{Quantitative comparison of flow-based methods with corrupted flow maps on the DAVIS 2016 validation set.}
\vspace{-2mm}
\small
\begin{tabular}{p{2.0cm}|P{0.7cm}|P{1.6cm}P{1.6cm}|P{0.7cm}}
\toprule
Method &Shift &$\mathcal{J}_\mathcal{M}$ &$\mathcal{F}_\mathcal{M}$ &Loss\\
\midrule
\multirow{3}*{MATNet~\cite{MATNet}} &0 &74.1 &77.1 &-\\
&16 &68.3 \textcolor{red}{(-5.8)} &68.8 \textcolor{red}{(-8.3)} &7.1\\
&32 &61.2 \textcolor{red}{(-12.9)} &61.1 \textcolor{red}{(-16.0)} &14.5\\
\midrule
\multirow{3}*{FSNet~\cite{FSNet}} &0 &83.2 &83.8 &-\\
&16 &70.9 \textcolor{red}{(-12.3)} &63.2 \textcolor{red}{(-20.6)} &16.5\\
&32 &60.0 \textcolor{red}{(-23.2)} &49.6 \textcolor{red}{(-34.2)} &28.7\\
\midrule
\multirow{3}*{AMC-Net~\cite{AMC-Net}} &0 &83.4 &84.9 &-\\
&16 &74.1 \textcolor{red}{(-9.3)} &72.1 \textcolor{red}{(-12.8)} &11.1\\
&32 &71.3 \textcolor{red}{(-12.1)} &69.4 \textcolor{red}{(-15.5)} &13.8\\
\midrule
\multirow{3}*{\textbf{TMO}} &0 &85.6 &86.6 &-\\
&16 &79.5 \textcolor{red}{(-6.1)} &80.3 \textcolor{red}{(-6.3)} &\underline{6.2}\\
&32 &77.8 \textcolor{red}{(-7.8)} &77.7 \textcolor{red}{(-8.9)} &\underline{8.4}\\
\midrule
\multirow{3}*{\textbf{TMO++}} &0 &85.6 &86.6 &-\\
&16 &81.1 \textcolor{red}{(-4.5)} &81.5 \textcolor{red}{(-5.1)} &\textbf{4.8}\\
&32 &80.4 \textcolor{red}{(-5.2)} &80.6 \textcolor{red}{(-6.0)} &\textbf{5.6}\\
\bottomrule
\end{tabular}
\label{table11}
\end{table}

\begin{figure*}[t]
\centering
\includegraphics[width=1\linewidth]{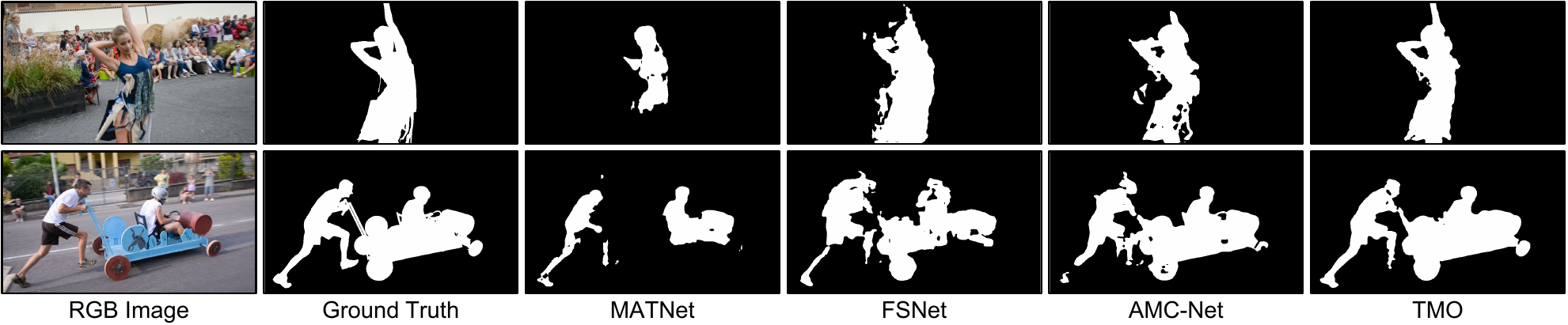}
\vspace{-6mm}
\caption{Qualitative comparison of flow-based approaches using corrupted flow maps.}
\label{figure7}
\end{figure*}

\vspace{1mm} 
\noindent\textbf{Robustness to noisy motion cues.}
A key strength of our approach lies in its robustness to low-quality motion cues during inference, which is achieved by minimizing reliance on motion branch. This is accomplished by randomly providing RGB images to the motion encoder, thereby reducing the influence of its embeddings during training. The adaptive output selection mechanism further enhances robustness by prioritizing more reliable predictions, ensuring resilience against noisy optical flow maps. We validate this through an experiment under noisy flow conditions, as shown in Table~\ref{table11}, where horizontal and vertical shifts simulate misalignment. The same noise generation process is applied to all methods, with pre-trained models for MATNet~\cite{MATNet} and FSNet~\cite{FSNet} from their official repositories, and AMC-Net~\cite{AMC-Net} trained from scratch.

The results demonstrate performance degradation as flow maps become misaligned, hindering accurate object detection. FSNet exhibits the largest performance drop due to its heavy reliance on motion cues. While MATNet and AMC-Net account for motion instability, their performance still declines. In contrast, our TMO maintains robustness by reducing reliance on the motion encoder and leveraging adaptive output selection to verify flow map reliability. Qualitative comparisons of predicted masks using corrupted flow maps, shown in Figure~\ref{figure7}, confirm that our method outperforms existing solutions, highlighting its reduced dependency on motion cues.

\begin{figure}[t!]
\centering
\includegraphics[width=1\linewidth]{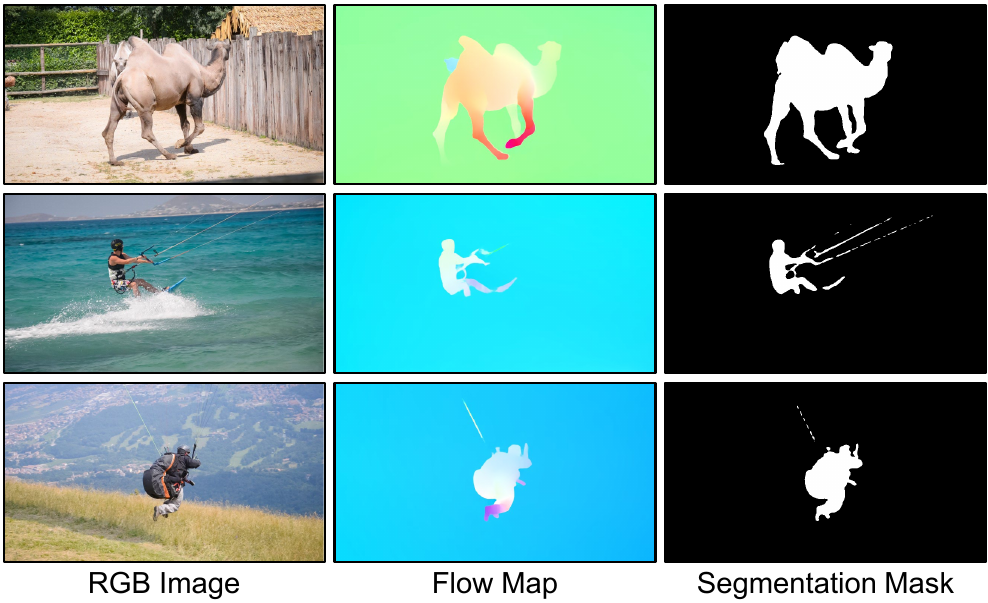}
\vspace{-6mm}
\caption{Visualized failure cases of our method due to visual vagueness.}
\label{figure8}
\end{figure}

\begin{figure*}[t!]
\centering
\includegraphics[width=1\linewidth]{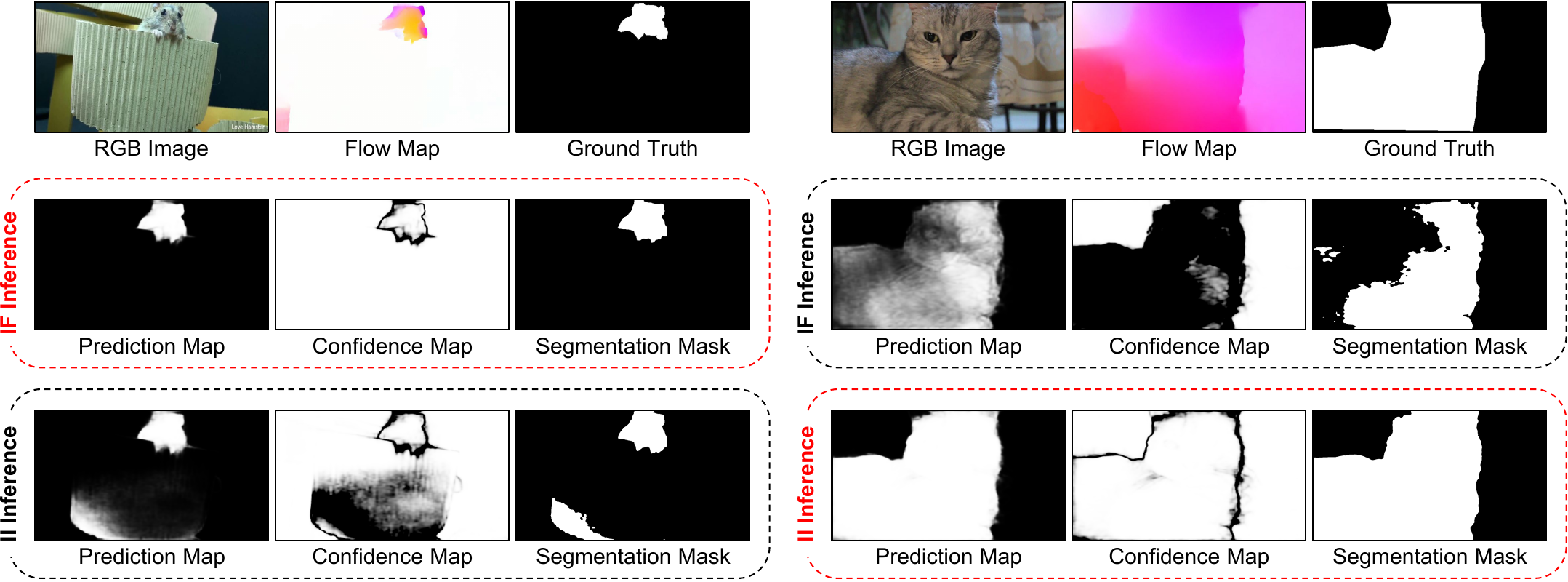}
\vspace{-6mm}
\caption{Visualized success cases of adaptive output selection, with the selected domain highlighted in red.}
\label{figure9}
\end{figure*}

\begin{figure*}[t!]
\centering
\includegraphics[width=1\linewidth]{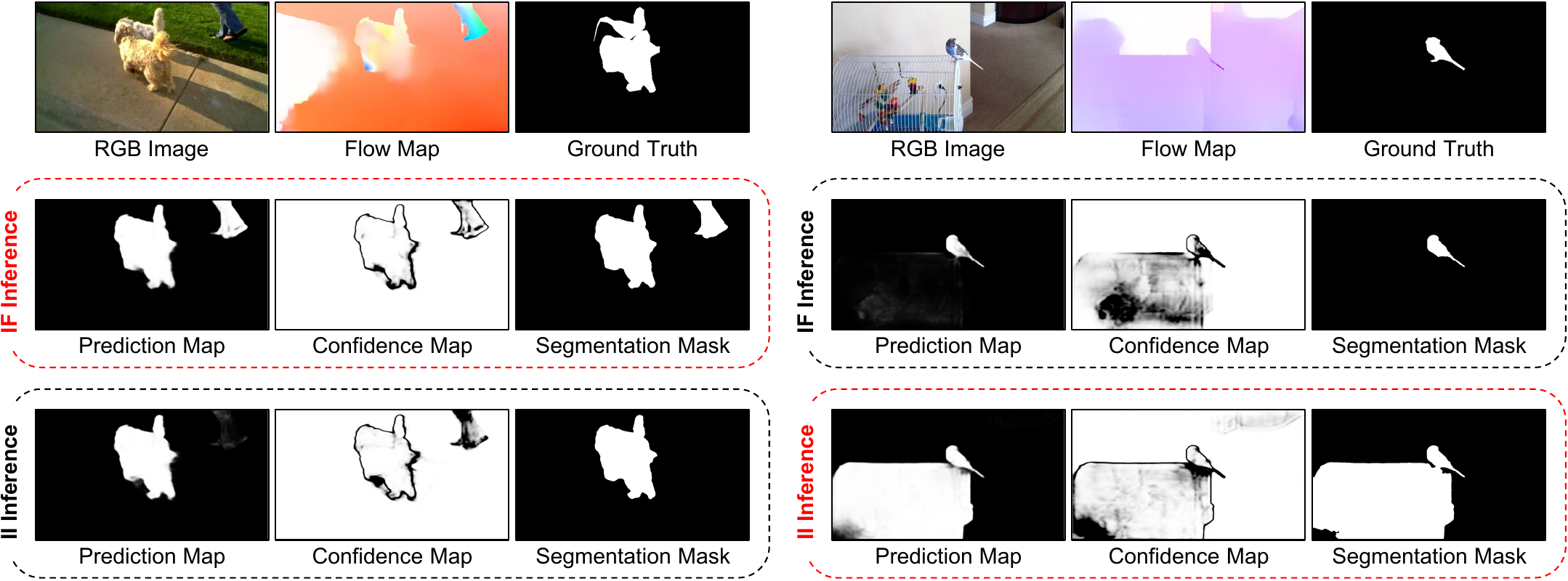}
\vspace{-6mm}
\caption{Visualized failure cases of adaptive output selection, with the selected domain highlighted in red.}
\label{figure10}
\end{figure*}

\section{Limitations}
\noindent\textbf{Visual vagueness.} 
Our method utilizes a two-stream pipeline that integrates appearance and motion cues to address visual vagueness. By employing a multi-modal fusion process, motion cues from optical flow maps complement RGB images, improving the model’s ability to segment objects in most cases. This combination helps address challenges posed by ambiguous or incomplete visual information. However, in rare scenarios where both appearance and motion cues are vague, the model struggles to reliably segment primary objects, leading to occasional inaccuracies.

Figure~\ref{figure8} illustrates these failure cases. In one example, the primary camel is connected to a background camel, creating significant ambiguity. The appearance encoder generates similar features for both camels, making motion cues essential for distinguishing the target object. Since the background camel also moves distinctly, the model mistakenly identifies it as part of the target. The other examples highlight difficulties with fine-grained details, such as thin structures. Optical flow maps fail to capture intricate features, leading to incomplete or inaccurate segmentation. These limitations emphasize the need for improvements in the model's ability to handle delicate structures or overlapping components.

To mitigate these issues, more advanced visual encoders and flow estimation models could better capture object-level representations and finer details, allowing for more precise segmentation. Additionally, specialized modules could be introduced to resolve spatial ambiguities and isolate objects in complex scenarios, thereby enhancing robustness.

\vspace{1mm}
\noindent\textbf{Inaccurate output selection.} 
Our adaptive output selection mechanism evaluates predictions from IF and II inference, selecting the most reliable output by comparing confidence scores. This approach assumes that higher confidence scores correspond to more accurate predictions, a principle that holds in most cases. By dynamically switching between inference modes based on confidence levels, the mechanism adapts to enhance segmentation quality.

As shown in Figure~\ref{figure9}, this assumption generally holds, with higher confidence scores aligning with better mask predictions. In one example, appearance cues fail to clearly indicate the target object, but motion cues from the optical flow map guide the prediction. IF inference, which uses motion cues, achieves higher confidence and more accurate segmentation. In contrast, weak motion values make motion cues unreliable in another case, and II inference, excluding motion cues, demonstrates higher confidence and a more precise mask. These examples highlight the mechanism’s ability to adaptively prioritize inference modes based on confidence levels, improving segmentation under challenging conditions.

However, as seen in Figure~\ref{figure10}, the mechanism can fail when confidence scores do not accurately reflect prediction quality. In one case, motion cues guide the detection of the person’s leg, resulting in higher confidence for IF inference. Background distractions cause overconfident but incorrect predictions. In another case, weak motion cues prompt the network to favor II inference, which shows higher confidence, but visual ambiguity leads to suboptimal output selection, producing inaccurate segmentation. These examples expose the limitations of the current approach.

While generally effective, the adaptive output selection mechanism has limitations due to its reliance on confidence scores as a proxy for prediction quality. To address these challenges, more robust evaluation methods, such as context-aware quality metrics, could be incorporated. Additionally, integrating uncertainty modeling could improve the mechanism's ability to assess the reliability of individual predictions, particularly in cases of conflicting or ambiguous cues. These enhancements would strengthen the selection process, enabling the network to handle complex scenarios with greater accuracy.

\section{Conclusion}
In unsupervised VOS, the combined use of appearance and motion cues has proven to be a powerful and effective approach. However, existing two-stream methods are often vulnerable to low-quality optical flow maps, which can compromise their usability and reliability in real-world scenarios. To address this issue, we propose a motion-as-option network that minimizes dependency on motion cues, coupled with a collaborative network learning strategy that fully exploits this unique property. Additionally, we introduce an adaptive output selection algorithm designed to optimize the performance of the motion-as-option network during inference. Our approach achieves new state-of-the-art performance across all public benchmark datasets while maintaining real-time inference speed. We believe that our simple, fast, and robust method can serve as a strong baseline for future VOS research.

\vspace{-1cm}
\begin{IEEEbiographynophoto}{Suhwan Cho}
received his B.S. and Ph.D degrees in Electrical and Electronic Engineering from Yonsei University, Seoul, Korea, in 2020 and 2025, respectively. He was a Research Scientist Intern in the Deep Learning Group at Adobe Research, San 
Jose, California, USA, in 2023. He is currently a Research Scientist at GenGenAI, Seoul, Korea. His current research interests include various aspects of video editing technologies, such as video object segmentation and video inpainting.
\end{IEEEbiographynophoto}

\vspace{-1.1cm}
\begin{IEEEbiographynophoto}{Minhyeok Lee} 
received his B.S. degree in Electrical and Electronic Engineering from Yonsei University, Seoul, Korea, in 2021, where he is currently pursuing a Ph.D. His current research interests include object segmentation, salient object detection, and vision-language retrieval. 
\end{IEEEbiographynophoto}

\vspace{-1.1cm}
\begin{IEEEbiographynophoto}{Jungho Lee} 
received his B.S. degree in Electrical and Electronic Engineering from Yonsei University, Seoul, Korea, in 2021, where he is currently pursuing a Ph.D. He was a Research Intern in the 3D Team at NAVER Cloud, Pangyo, Korea, in 2024. His current research interests include human motion analysis and neural view synthesis. 
\end{IEEEbiographynophoto}

\vspace{-1.1cm}
\begin{IEEEbiographynophoto}{MyeongAh Cho} 
received her B.S. degree in Electronic Engineering from Kyung Hee University, Yongin, Korea, in 2018. She received her Ph.D. degree in Electrical and Electronic Engineering from Yonsei University, Seoul, Korea, in 2023. She was a Research Intern in the Video Team at NAVER CLOVA, Pangyo, Korea, in 2023. She is currently an Assistant Professor in the Department of Software Convergence, Kyung Hee University, Yongin, Korea. Her current research interests include video anomaly detection and face recognition.
\end{IEEEbiographynophoto}

\vspace{-1.1cm}
\begin{IEEEbiographynophoto}{Seungwook Park} 
received his B.S. degree in Electrical and Electronic Engineering from Korea University, Seoul, Korea, in 2007. He is currently with the EO/IR Systems R\&D Lab at LIG Nex1. His research interests include EO/IR systems and large-aperture telescope systems.
\end{IEEEbiographynophoto}

\vspace{-1.1cm}
\begin{IEEEbiographynophoto}{Jaeyeob Kim} 
received his M.S. degree in Electronic Engineering from Hanyang University, Seoul, Korea, in 2022. He is currently with the EO/IR Systems R\&D Lab at LIG Nex1. His research interests include image deblurring, super-resolution, and object tracking in EO/IR systems.
\end{IEEEbiographynophoto}

\vspace{-1.1cm}
\begin{IEEEbiographynophoto}{Hyunsung Jang} received his B.S. and M.S. degrees in Electrical and Computer Engineering from Pusan National University, Busan, Korea, in 2007 and 2009, respectively. He is currently pursuing a Ph.D. in Electrical and Electronic Engineering at Yonsei University, Seoul, Korea. He serves as a Chief Research Engineer in the EO/IR Systems R\&D Lab at LIG Nex1. His research interests include computer vision and machine learning.
\end{IEEEbiographynophoto}

\vspace{-1.1cm}
\begin{IEEEbiographynophoto}{Sangyoun Lee} (\textit{Member, IEEE}) 
received his B.S. and M.S. degrees in Electrical and Electronic Engineering from Yonsei University, Seoul, Korea, in 1987 and 1989, respectively. He received his Ph.D. degree in Electrical and Computer Engineering from the Georgia Institute of Technology, Atlanta, Georgia, USA, in 1999. He is currently a Professor at the School of Electrical and Electronic Engineering, Yonsei University, Seoul, Korea. His research interests include all aspects of computer vision.
\end{IEEEbiographynophoto}

\vfill
\end{document}